%% file: ijcai22.tex
\documentclass{article}
\usepackage{ijcai22}

\usepackage{times}
\usepackage{soul}
\usepackage{url}
\usepackage[hidelinks]{hyperref}
\usepackage[utf8]{inputenc}
\usepackage[small]{caption}
\usepackage{graphicx}
\usepackage{amsmath}
\usepackage{amsthm}
\usepackage{booktabs}
\usepackage{algorithm}
\usepackage{algorithmic}
\title{Mediators: \\ Conversational Agents Explaining NLP Model Behavior}

\author{
    Nils Feldhus$^1$
    \and
    Ajay Madhavan Ravichandran$^{1,2}$
    \And
    Sebastian M\"oller$^{1,2}$
    \affiliations
    $^1$German Research Center for Artificial Intelligence (DFKI) \\
    $^2$Technische Universit\"at Berlin \\
    \emails
    \{nils.feldhus, ajay\_madhavan.ravichandran,
    sebastian.moeller\}@dfki.de
}

\begin{document}
\input{main}

%% The file named.bst is a bibliography style file for BibTeX 0.99c
\bibliographystyle{named}
\bibliography{main}
\end{document}

%% file: main.tex
\maketitle

\begin{abstract}
\input{sections/0_abstract}
\end{abstract}

\input{sections/1_introduction}

\input{sections/2_motivation}
%%%%%%%%%%
\input{figures/sst_example}
%%%%%%%%%%
\input{sections/3_generating}

\input{sections/4_responding}
%%%%%%%%%%
\input{figures/follow-up}
%%%%%%%%%%
\input{sections/5_understanding}

\input{sections/6_state-tracking}

\input{sections/7_beyond}

\input{sections/8_training}
\input{sections/9_related-work}
\input{sections/10_conclusion}
\input{sections/acknowledgments}

%% file: sections/0_abstract.tex
The human-centric explainable artificial intelligence (HCXAI) community has raised the need for framing the explanation process as a conversation between human and machine. 
In this position paper, we establish desiderata for Mediators, text-based conversational agents which are capable of explaining the behavior of neural models interactively using natural language.
From the perspective of natural language processing (NLP) research, we engineer a blueprint of such a Mediator for the task of sentiment analysis and assess how far along current research is on the path towards dialogue-based explanations.

%% file: sections/1_introduction.tex
\section{Introduction}

In almost all areas of artificial intelligence, there is continuous evidence that neural models with an ever-growing number of parameters and training data are here to stay, thanks to scaling laws \cite{kaplan-2020-scaling-laws}.
Explaining the behavior of these large models has taken the center stage in many areas including NLP research \cite{madsen-2021-post-hoc}.
Interactivity in explainable artificial intelligence (XAI) has been a hot topic for a while \cite{abdul-2018-trends} and framing the explanation process as a dialogue between the human and the model has solid theoretical foundations in the HCXAI literature \cite{miller-2019-explanation,weld-bansal-2019-crafting,liao-varshney-2021-hcxai,lakkaraju-2022-rethinking,dazeley-2021-levels,mariotti-2020-harnessing},
but no prior work has specified how to apply these frameworks to NLP problems and language models.
Simultaneously, a big push from the NLP community towards implementing such systems has yet to occur.

\input{figures/mediator_concept}

In this paper, we highlight three key factors motivating the use of conversational agents to explain the behavior of NLP models (\S \ref{sec:motivation}):
The flexibility of natural language, the need for a complementary view for explainability methods, and the need to alleviate the cognitive load from the explainee.

Following \cite{sokol-flach-2020-one} and \cite{lakkaraju-2022-rethinking}, we envision such conversational agents as complex, modular systems that we coin \textit{Mediators}.
Conceptually, a Mediator has to generate appropriate atomic explanations (\S \ref{sec:generating}), 
respond to the user in natural language (\S \ref{sec:responding}), 
understand a user's natural language input (\S \ref{sec:understanding}),
and keep track of the conversation and the user's knowledge (\S \ref{sec:state}).
These tasks are part of a process that we depict in Fig.~\ref{fig:concept}. %, including the initial input and model prediction as well as dialogue feedback to improve the explained model.
For the scope of this paper, we focus on purely textual setups and NLP models as the explanandum, but the modular structure of Mediators applies to other types of black boxes and multimodal interactive frameworks \cite{voigt-2021-designing} as well.

We engineer a blueprint of a Mediator for NLP model behavior by taking its modules apart and devising examples for the downstream task of sentiment analysis% and contextualizing each module in terms of the current progress on its implementation
.
Simultaneously, we extend the seminal work of \cite{miller-2019-explanation} and the recent works of \cite{nobani-2021-conversational} and \cite{lakkaraju-2022-rethinking} by %recommending best practices for modeling Mediators and deploying them in real-world settings.
raising awareness of further aspects to consider such as evaluation and customization of the conversation as well as data collection (\S \ref{sec:beyond}). 
We hope that this position paper aiming at NLP practitioners helps in closing the gap of conversational explainability.

%% file: figures/mediator_concept.tex
\begin{figure}[t!]
    \centering
    \resizebox{\columnwidth}{!}{%
    
    \includegraphics{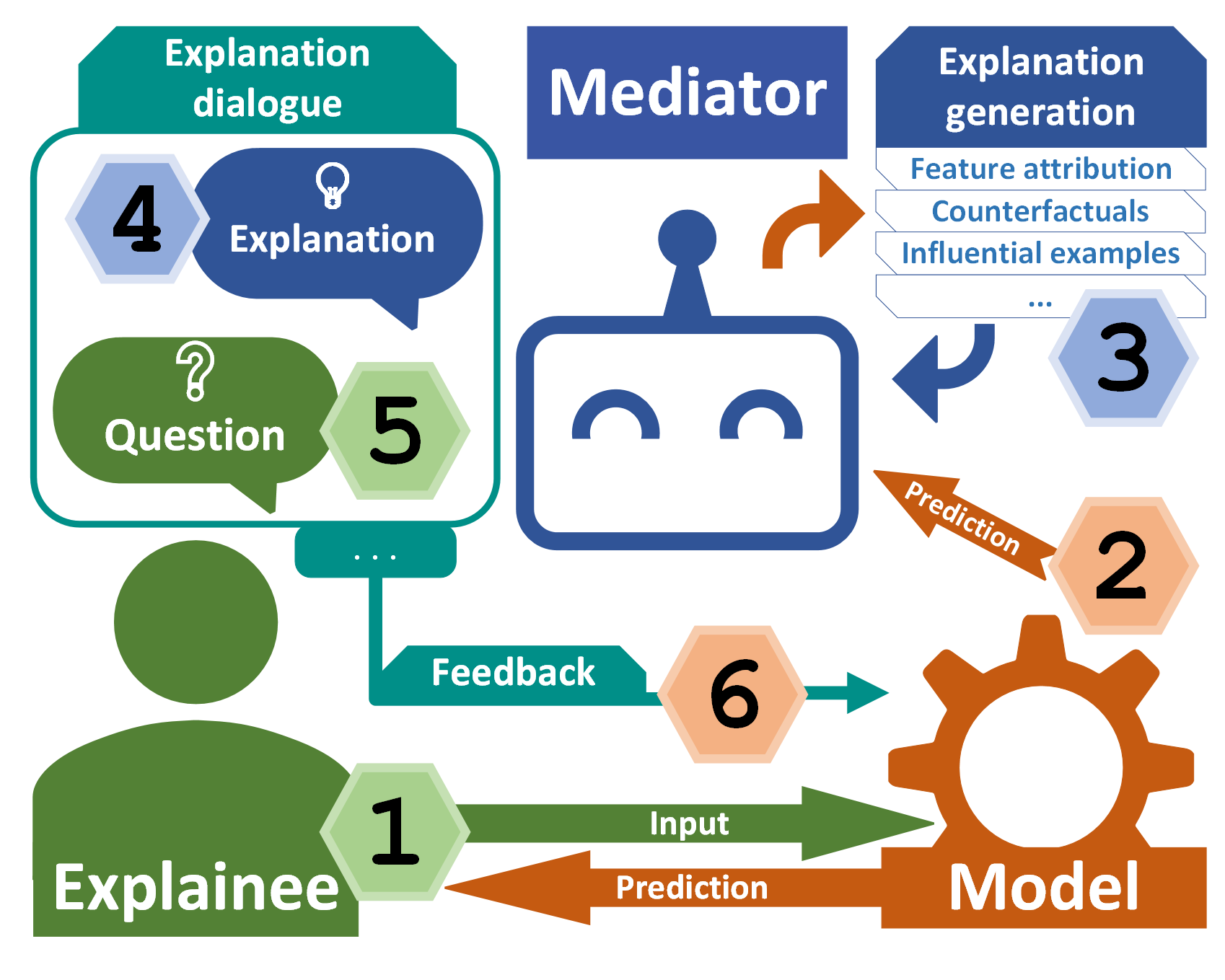}
    
    }
    %\vspace*{-5mm}
    \caption{Simplified concept of a Mediator explaining the predictions of a Model to the human Explainee. \\
    \raisebox{.5pt}{\textcircled{\raisebox{-.9pt} {1}}}: The Explainee provides input to the Model. \\ \raisebox{.5pt}{\textcircled{\raisebox{-.9pt} {2}}}: The Model puts out a prediction based on the input. \\ \raisebox{.5pt}{\textcircled{\raisebox{-.9pt} {3}}}: The Mediator generates explanations based on the prediction. \\ \raisebox{.5pt}{\textcircled{\raisebox{-.9pt} {4}}}: The Mediator holds an explanation dialogue with the Explainee. \\ \raisebox{.5pt}{\textcircled{\raisebox{-.9pt} {5}}}: The Explainee acts upon the explanation and asks follow-up questions. \raisebox{.5pt}{\textcircled{\raisebox{-.9pt} {4}}} and \raisebox{.5pt}{\textcircled{\raisebox{-.9pt} {5}}} are repeated until the Explainee is satisfied. \\ %The Mediator keeps track of the dialogue state and the user's mental model.
    \raisebox{.5pt}{\textcircled{\raisebox{-.9pt} {6}}}: The explanation dialogue serves as corrective feedback which the Model can improve on. }
    %\caption{Simplified concept of a Mediator explaining the predictions of a Model to the human Explainee: The model puts out a prediction based on the input from the Explainee. The Mediator generates atomic explanations based on the prediction and question from the Explainee in order to hold an explanation dialogue with the Explainee. The Explainee asks follow-up questions until satisfied.}
    \label{fig:concept}
\end{figure}

%% file: sections/2_motivation.tex
\section{Why we need Mediators}
\label{sec:motivation}

\subsection{A complementary view of explainability methods}

Natural language is said to be the most accessible and human-centric modality of explanation \cite{ehsan-riedl-2020-hcxai}.
Natural language explanations also exceed other explainability methods in plausibility, i.e. how convincing they are to the human explainee \cite{lei-2016-rationalizing,lipton-2018-mythos,camburu-2018-esnli,jacovi-goldberg-2020-towards}.

Natural language explanations alone, however, might cause unwarranted trust from the human recipient \cite{jacovi-2021-formalizing-trust}, because the models which generate them usually are optimized via direct supervision towards a few human-acceptable gold rationales from static datasets \cite{camburu-2018-esnli}.
Other explainability methods excel in faithfulness (sometimes called fidelity) \cite{jacovi-goldberg-2020-towards}, measuring how truthful an explanation reflects the internal representations of the explained model.
In particular, \cite{wiegreffe-2021-measuring} found that faithful explanations are not achievable with free-text rationalization models and truly faithful explanations might be impossible, after all \cite{jacovi-goldberg-2020-towards}.

That is why a complementary view of explainability methods (for NLP) \cite{madsen-2021-post-hoc,jacovi-2022-diagnosing} is necessary, where faithful and plausible \cite{jacovi-goldberg-2020-towards,herman-2017-promise,gilpin-2018-explaining} explanations with a varying scope can both be presented to a user.
This has previously been advocated for by works in HCXAI such as \cite{hohman-2019-gamut} and \cite{yeung-2020-sequential} among others.
Since people are better at understanding narratives rather than numbers and probabilities, \cite{reiter-2019-natural} proposed to present the reasoning done by a numerical non-symbolic model (such as a large neural model) as a narrative including causal and argumentative relations.

\subsection{Cognitive load}

However, next to explanations being faithful and plausible, 
we also detect a need for them to be both sufficient and concise: 
% MH: Is this the same as the soundness vs completeness trade-off mentioned here: https://aclanthology.org/W18-6511.pdf (Sec 5)?
% NF: No, the completeness measure in that paper refers to decision trees.
\cite{deyoung-2020-eraser} define sufficiency as a measure evaluating if the amount of information communicated through an explanation is enough to justify some model predictions. 
While it might be tempting to densely pack as many cognitive chunks \cite{doshi-velez-kim-2017-rigorous} into an explanation,
it may not be a good explanation in terms of overall satisfaction \cite{lombrozo-2007-simplicity}, understandability \cite{ehsan-2019-automated-rg} and usefulness \cite{bansal-2021-whole}.

%%% Cognitive load, selection, conversation
The natural process for human explainers is to
select only the most relevant causes for an event. Although an event's causal chain often is much longer, humans have to make conscious decisions about what to say, since the cognitive load on the recipient's end might become too big otherwise \cite{hilton-2017-social,miller-2019-explanation,doshi-velez-kim-2017-rigorous}.
Systems implemented in related work are prone to overwhelm users
with too much information at once \cite{kulesza-2015-elucidebug}.
Natural language can help here, because it is flexible, i.e. it can be tuned to custom amounts of cognitive chunks and can be adapted to both different audiences and tasks.

\subsection{The case for dialogue-based explanations}
\label{par:case-dialogue}

The aforementioned cognitive limits motivate a setup where concise explanations are presented to the user at each turn prompting them to ask follow-up questions until their needs for information are satisfied \cite{leake-1991-goal-based}, thus engaging in back-and-forth conversations \cite{weld-bansal-2019-crafting,liao-varshney-2021-hcxai,madumal-2019-grounded,hartmann-2021-interaction,sevastjanova-2018-going}.
These concise explanations should be ``atomic'' in nature and contain only a few cognitive chunks to not overwhelm the user.

User satisfaction has also been shown to increase under the presence of an option to provide feedback. In particular, \cite{smith-renner-2020-accountability} concluded that it reduces user frustration and leads them to understand the model in the background better.
The findings of \cite{lakkaraju-2022-rethinking} indicate that decision-makers would strongly prefer interactions to take the form of natural language dialogues, in order to treat machine learning models as ``another colleague'' who can be held accountable by asking why they made a particular decision through expressive and accessible natural language interactions.

In summary, we argue that by drawing from many different types of explainability methods and presenting atomic explanations on the basis of cognitive chunks, conversational agents with a modular setup like Mediators can be employed to interact with users in a natural language dialogue.

%% file: figures/sst_example.tex
\begin{figure*}
\centering
\textbf{Input} $x :=$ \textit{\textbf{\ul{the year 's best and most unpredictable comedy}}} $\Rightarrow$ \textbf{Model prediction} $y =$ \textbf{\ul{positive}} \\
\end{figure*}
\smallskip

\begin{table*}[ht!]
    \centering
    \renewcommand{\arraystretch}{1.5}
    
    \resizebox{\textwidth}{!}{%
    
    \begin{tabular}{r|l|l}
    \toprule 
    \textbf{Method}
    & \textbf{Question / User utterance}
    & \textbf{Explanation / Response from Mediator} \\
    
    \midrule
    Feature Attribution
    & Which tokens are most important for the prediction?
    & \textit{\ul{best}} and \textit{\ul{unpredictable}} are most important. \\
    
    Adversarial Examples
    & What would break the model's prediction?
    & Changing \textit{\ul{best}} to \textit{\ul{finest}} flips it to \ul{negative}. \\
    
    Influential Examples
    & What training examples influenced the prediction?
    & \textit{\ul{a delightfully unpredictable , hilarious comedy}} \\
    & & is an influential instance also classified as positive. \\
    
    Counterfactuals
    & What does the model consider a valid opposite example? 
    & Changing \textit{\ul{best}} to \textit{\ul{worst}} and \textit{\ul{unpredictable}} \\
    & & to \textit{\ul{predictable}} creates a valid \ul{negative} instance. \\
    
    Model Rationales
    & What would a generated natural language explanation be?
    & \textit{\ul{Unpredictable comedies are funny}}. \\
    
    \bottomrule
    \end{tabular} }
    \caption{Example instance $x$ from the SST dataset \protect\cite{socher-2013-recursive} that an explainee uses as input to a language model tackling the task of sentiment analysis (distinguishing positive from negative movie reviews). The table depicts the taxonomy of explainability methods and their associated questions as they appear in \protect\cite{madsen-2021-post-hoc}. (Please consult their paper for more information on available methods.) In the right column, we added explanation candidates that could serve as answers presented to the explainee by the Mediator. The underlined parts are the actual output of the respective method, while the rest is generated through verbalization. 
    }
    \label{tab:sst}
\end{table*}

%% file: sections/3_generating.tex
\section{Generating atomic explanations}
\label{sec:generating}

\paragraph{Explanations should not be static}
As \cite{hohman-2019-gamut} and \cite{lakkaraju-2022-rethinking} pointed out,
most of the existing work on explainability focuses on one-off, static explanations, e.g. feature attribution methods producing a single saliency map \footnote{In NLP, these explanations are commonly produced with libraries such as AllenNLP Interpret \cite{wallace-2019-allennlp} and Captum \cite{kokhlikyan-2020-captum} which can easily be applied to the wide array of pre-trained language models contributed through, e.g., Hugging Face transformers \cite{wolf-2020-transformers}, as demonstrated in Thermostat \cite{feldhus-2021-thermostat}.}
that conveys a limited amount of information and nothing about the causality involved.
Although the explainable NLP community has come up with text- and task-specific methods in the last few years \cite{madsen-2021-post-hoc}, there is still no apparent push towards conversation-based explanations of NLP model behavior.

\paragraph{Explanation generation as a selection process}
\cite{miller-2019-explanation} looked into the processes of how people select explanations from available causes, by following common heuristics such as abnormality, intentionality, necessity, sufficiency, and robustness.
This is underlined by the notion that there is no single best explanation \cite{ribera-lapedriza-2019-better}. By framing explanation generation as a search or exploration problem, we can incorporate many factors, including those correlating with human preferences, to find an optimal candidate.
This refers to both (1) selecting the best explanation among explanations of the same type, e.g. counterfactuals, and (2) selecting the best explanation type, e.g. a saliency map vs. a counterfactual.

The way explanations are commonly employed may not be sufficient for exploration and continuous discovery from users that have a range of skills and expertise.
In this regard, \cite{sokol-flach-2020-one} highlight that explanations have properties such as context, scope and breadth that may have to be personalized when generating explanations.
Being able to navigate between different kinds of explanations was also advocated for by \cite{bove-2022-contextualization} who developed an interface for contextualizing feature attribution explanations. We argue that such an exploration can also be modeled as a conversation.
One specific implementation of a search-based explanation generation is the work of \cite{wiegreffe-2022-reframing} whose framework over-generates explanation candidates using a language model and subsequently filters them using a second language model that is trained on human acceptability ratings collected in a crowd study.
\cite{hase-2021-ood} devised methods for searching through the space of possible explanations as an alternative to existing feature attribution methods.
\cite{treviso-martins-2020-explanation} framed explainability as a communication problem between an explainer and a layperson about a model's decision and empirically assessed the quality of the explanation. %They implemented sparse, selective attention mechanisms to maximize the success rate.

We argue that the methods mentioned and categorized in \cite{jacovi-2022-diagnosing} and \cite{madsen-2021-post-hoc} could be used in a complementary view of explainability for neural models: Feature attribution, counterfactuals, influence functions, and model-generated rationales often occur separately, but a Mediator has to be able to draw upon this pool of explanations generated from different methods dynamically based on user needs.\footnote{This task is depicted as Step 3 in Fig.~\ref{fig:concept}.}
In Tab.~\ref{tab:sst}, we showcase the example use case of sentiment analysis, a binary text classification task, and devise questions associated to explainability methods from \cite{madsen-2021-post-hoc}.\footnote{We identify verbalizing explanations, i.e. translating them into natural language, such as feature attribution and adversarial examples as a missing component. Existing solutions include the works of \cite{forrest-2018-towards} and \cite{yao-2021-refining}.}
Feature attribution explanations and model rationales can serve as initial explanations letting explainees form their hypotheses about the model's behavior, while counterfactuals and adversarial examples can be sanity checks that support or counter the hypotheses \cite{hohman-2019-gamut}.

Sentiment analysis, like most text classification tasks, is a trivial task in terms of explanation generation. However, since other, more challenging tasks are less explored for most types of explainability methods, we find it fitting for the scope of our work. We identify methods for explaining text generation or, more generally, language modeling \cite{vafa-2021-greedy,yin-neubig-2022-interpreting} as a very promising avenue for future research. This opens up the pathway towards real-world use cases such as question answering and machine translation.

%% file: sections/4_responding.tex
\section{Responding to the user in natural language}
\label{sec:responding}

After collecting sufficiently many explanations for a Mediator to choose from, the next hurdle is to verbalize and present them in a way that engages the user to start, continue and finish a conversation.\footnote{This task is depicted as Step 4 in Fig.~\ref{fig:concept}.}
We understand explanations to take the form of information-seeking dialogues \cite{walton-krabbe-1995-commitment}, where the user seeks the answers to some questions and the Mediator knows and provides them.
In the case of explanation dialogues, we identify a mixed initiative setting. The start of the conversation can either be triggered by the Mediator presenting a concise explanation that prompts the user to interact with it, or by the human who already has a clear goal, e.g. an explanation type, in mind.
Separating the explanation content planning from the execution of the dialogue has been proposed as early as \cite{cawsey-1991-interactive}.
The Mediator should respond with informative and properly contextualized explanations for why the underlying model made specific decisions \cite{lakkaraju-2022-rethinking}.
%There is a line of research using computational argumentation for dialogue-based explanations \cite{antaki-leudar-1992-argument,vassiliades-2021-argumentation}. This has yet to be explored in depth, as to how this can be applied to our proposal for Mediators.
%the main purpose of Mediators is to transfer knowledge and build a mental model about model behavior instead of persuading the user about a fixed statement.
%Both the model and the Mediator are unlikely to convey the true underlying causes for some prediction or outcome, because true faithfulness might be impossible to achieve, and even then such an explanation will not be plausible \cite{jacovi-2022-diagnosing}.
%We see the next best solution is to provide a plausible explanation that is concise, sufficiently informative and relevant to their utterance.
%\paragraph{Personalized responses}
According to the maxims of relation and quantity, it is essential to only relay information that is relevant and necessary at any given point in time. It means that a Mediator has to ``know'' what the user knows and expects (\S \ref{sec:beyond}) before determining the content of the explanation \cite{sokol-flach-2020-fact-sheets,hartmann-2021-interaction}.
At the same time, in accordance with the notion of parsimony \cite{sokol-flach-2020-fact-sheets}, Mediators should aim to fill in the most gaps \cite{leake-1991-goal-based} with the fewest statements.
This ties back in with the issue of cognitive load: Although the frame of a conversation alleviates this issue on a high level, every single response to the user should be selected with user knowledge and expectations in mind.

\cite{akula-2019-natural} provided answers to user questions using different reasoning paradigms in a visual setting.
\cite{madumal-2019-grounded} developed a framework that allows users to follow up on an explanation to reach a comprehensive model understanding.

We urge researchers to investigate different approaches regarding explanation selection and generation: While we argue for a pool of explanations produced by various methods that the Mediator can draw from (\S \ref{sec:generating}), one might also train a framework in an end-to-end fashion combining selection, generation and responding. 

%% file: figures/follow-up.tex
\begin{table}[t!]
    \centering

    \resizebox{\columnwidth}{!}{%
    
    \begin{tabular}{r|l}
    \toprule
    \textbf{Category of} \\
    \textbf{follow-up question} & \textbf{Example questions} \\
    
    \midrule
    Input text edits
    & What if we removed word $w$ from the input? \\
    & What if we added the phrase $p$ at the end? \\
    & What if the sentence $s$ was in passive voice? \\
    
    \midrule
    Scope restrictions
    & What change in the phrase $p$ \\
    & \quad would flip the prediction? [AE] \\
    & How does word $w$ need to be changed \\
    & \quad in order to flip the prediction? [AE] \\
    & What is the most salient word \\
    & \quad in the $n$-th sentence? [FA] \\
    
    \midrule
    Foil edits
    & What are the most salient tokens \\
    & \quad for class $y'$ instead of $y$? [FA] \\
    & What training example from class $y'$ \\
    & \quad influenced the prediction the most? [IE] \\
    
    \midrule
    Explanation
    & Could you show me the LIME instead of \\
    source edits
    & \quad the Shapley Values explanation? [FA] \\
    & Could you show me the Integrated Gradients \\
    & \quad explanation with 50 samples? [FA] \\
    
    \bottomrule
    \end{tabular} }
    \caption{Categories and examples of follow-up questions beyond the five generic questions in Tab.~\ref{tab:sst} that can trivially be mapped to an explanation type. Associated explanation type (FA: Feature Attribution, AE: Adversarial Examples, IE: Influential Examples) in square brackets (none means the question is applicable to any type).
    All except ``explanation source edits'' were already proposed in \protect\cite{weld-bansal-2019-crafting}.
    }
    \label{tab:follow-up}
\end{table}

%% file: sections/5_understanding.tex
\section{Understanding a user's natural language input}
\label{sec:understanding}

A Mediator should understand continuous requests for explanations and be able to efficiently map these to appropriate explanation types to generate \cite{lakkaraju-2022-rethinking}.
This is commonly understood as an intent recognition problem in dialogue and would be realized as one module in a Mediator framework.\footnote{This corresponds to Step 5 in Fig.~\ref{fig:concept}.}
\cite{lakkaraju-2022-rethinking} identified natural language understanding as a problem, because there is a large set of possible query types, many different ways to phrase explainability questions, and transferring it to different downstream applications is generally very complex.
\cite{weld-bansal-2019-crafting} outlined the types of follow-up and drill-down actions a user might request upon seeing some initial explanation.

For this module that is tasked with recognizing the user intent, the canvas of \cite{lim-dey-2009-demand} -- later expanded upon by \cite{weld-bansal-2019-crafting} and \cite{liao-2021-question-driven} -- can serve as a general-purpose mapping between user utterance and explanation type.
We found \cite{rebanal-2021-xalgo} to be the only work that concretely built a classifier for user questions. Based on the classification, they generated explanations for the learned representations of the underlying model. 
We see the taxonomy of \cite{madsen-2021-post-hoc} (Tab.~\ref{tab:sst}) as a starting point in adapting such a mapping to NLP settings and propose four distinct types of follow-up questions in Tab.~\ref{tab:follow-up}: Input text edits, scope restrictions and foil edits which were proposed in \cite{weld-bansal-2019-crafting} as well as explanation source edits which is more targeted at model developers and ML experts rather than laypeople.
The challenge then is to tie the intent recognition module to the search method that identifies potential explanation candidates and optionally combine them to end up with the final answer presented to the explainee.
In some cases, e.g. ``What if the sentence $s$ was in passive voice?'', the module has to have a thorough understanding of language and perform edits accordingly.

%% file: sections/6_state-tracking.tex
\section{Keeping track of the dialogue}
\label{sec:state}

When designing dialogue systems, the task of keeping track of the dialogue history is essential to better inform the selection of the next action or response.
This is traditionally done by predicting slot-value pairs representing the user's goals while accounting for the dialogue history at any turn.
We point the reader towards recent methods in dialogue state tracking \cite{balaraman-2021-dst,hu-2022-in-context} and dialogue systems for information acquisition \cite{cai-2022-learning}.
All of them are disconnected from the explainability literature, however.

\input{figures/dialogue}

\paragraph{User models and mental models}
The explainability literature, on the other hand, has explored user models to keep track of the recipient's knowledge.
The idea of a user model has been explored in Cawsey's EDGE system \shortcite{cawsey-1991-interactive}: The knowledge that the user has about a phenomenon and their level of expertise should both be updated during the dialogue \cite{miller-2019-explanation}.
\cite{stumpf-2009-interacting} allowed users to interact with different types of explanations and examined if these enable them to form useful mental models of the system. In a follow-up work, \cite{kulesza-2015-elucidebug} presented explanatory debugging as a use case to address this question about mental models, allowing users to communicate corrections back to the system.
\cite{weld-bansal-2019-crafting} pointed out that constructing user models, i.e. tracking explicitly what users know and expect, are typically based on hand-engineered solutions.
In the remainder of this section, we will highlight different aspects related to user models and mental models. We illustrate this relation in Fig.~\ref{fig:dialogue}.

\paragraph{Addressing misalignments between model and user expectation}
Model predictions, the output of explainability methods and user expectations are often misaligned \cite{schuff-2022-human}. 
This leads us to the analysis of the users' mental models: Keeping track of the dialogue also means estimating the users' understanding of the underlying model's behavior, e.g. by using a formal argumentative dialogue framework \cite{madumal-2019-grounded,sokol-flach-2020-one} or by using simulatability \cite{doshi-velez-kim-2017-rigorous,hase-bansal-2020-evaluating} tests prompting them to simulate the model on unseen data.
In practice, the latter would usually occur in a separate evaluation stage after users have interacted with the model, but a rigorous evaluation would require a frequent assessment of the user's understanding.

Another avenue relevant to these misalignments is the one of active learning \cite{ghai-2020-xal}. For example, the workflow by \cite{liang-2020-alice} presents most confusing class pairs to human experts and queries them for explanations. Such a system can then utilize this new knowledge to improve the underlying classification model.
Hence, an improved underlying model can also mitigate the mismatch between user and model expectations. Two further recent works dealing with learning from natural language explanations have specified empirically which type of explanation data helps models the most in terms of task performance \cite{hase-bansal-2022-roles} and which strategies for training on such data are preferable \cite{carton-2022-effective-learning}.

\paragraph{User expertise}
\cite{ehsan-2021-who} examined and discussed the effect of explanations on groups of users with different AI backgrounds. Elaborate Mediator designs need to take into account that there might not be a one-for-all solution \cite{sokol-flach-2020-one} even when a user's mental model is considered \cite{chromik-2021-point}.
We think the field will eventually catch up more with targeting laypeople as recipients for explanations, which would make this aspect even more relevant than it already is.

\paragraph{Reacting to user feedback}

\cite{miller-2019-explanation} posed the question: ``what should an explanatory agent do if the explainee does not accept a selected explanation?''
We argue that all user feedback should be considered as training signals for (a) the underlying (explained) model, (b) the explanation-generating model, and (c) the user model that keeps track of user knowledge and the dialogue.
Thus, most kind of user feedback to Mediators is \textit{rich feedback} which involves ``more expressive forms of corrective feedback which can cause a deeper change in the underlying machine learning algorithm'' \cite{stumpf-2009-interacting}. We attenuate this slightly, because costly model edits or retraining may impede the user experience. It might even degrade the model performance due to misconceptions and biases of the explainee. In addition, incomplete statements such as ``The model made a mistake.'' or ``This explanation is wrong.'' should not be taken into account.
The advisory dialogues of \cite{moore-paris-1993-planning} explicitly modeled the effect of utterances on the recipient's mental state allowing for a recovery mechanism from failure and misunderstanding \cite{miller-2019-explanation}. \footnote{Fig.~\ref{fig:concept} depicts this feedback with Step 6.}

\paragraph{Finishing an explanation}
\cite{miller-2019-explanation} posed the question of ``how do we know that an explanation has 'finished'?'' which can involve knowing whether the explainee has correctly understood the explanation.
\cite{alvarez-melis-2019-weight} proposed a Weight-of-Evidence metric measuring the effect of ``explaining away'' different outcomes. From a theoretic perspective, it might suffice to exhaust all alternatives, but cannot be guaranteed to fulfill the user's understanding in practice.
Therefore, we highlight this as a potential roadblock.

%% file: figures/dialogue.tex
\begin{figure}[ht!]
    \centering
    \resizebox{\columnwidth}{!}{%
    
    \includegraphics{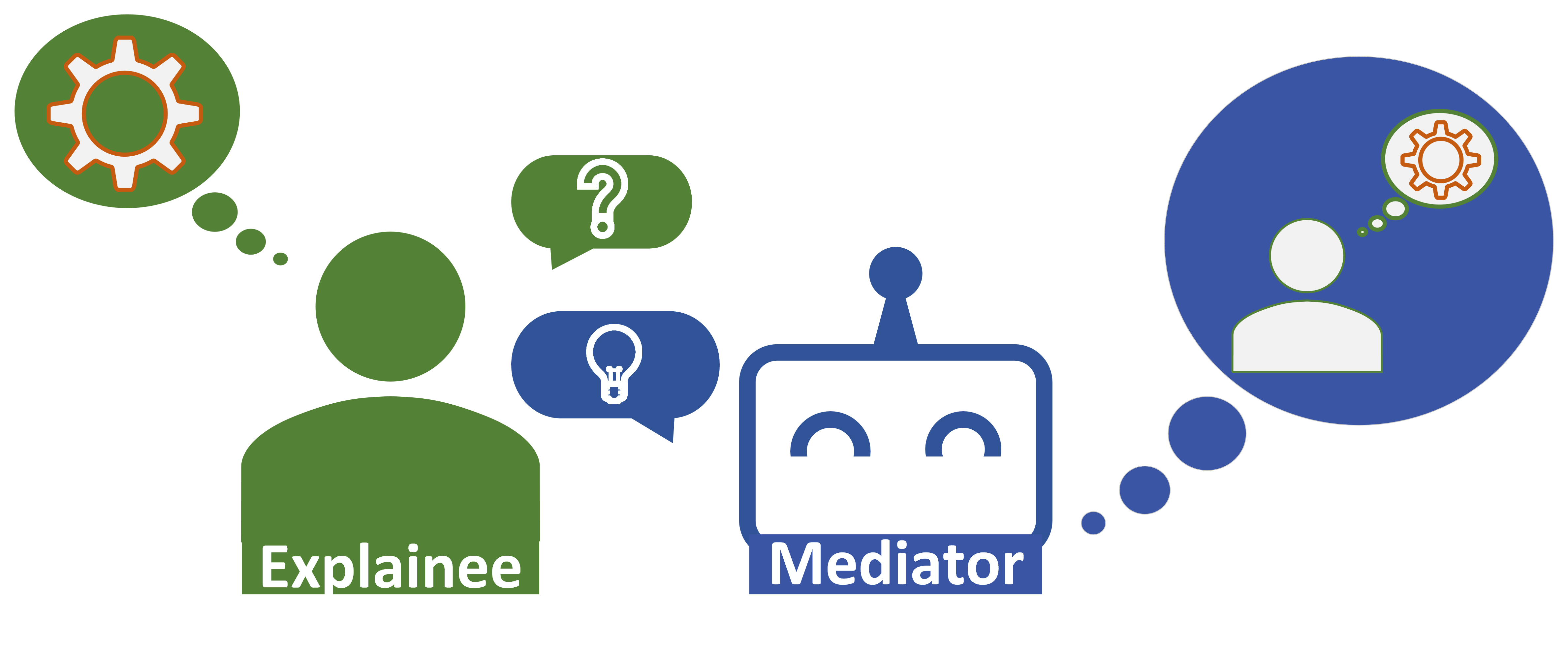}
    
    }
    \caption{Depiction of the Explainee's mental model of the explained Model and the Mediator's model of the Explainee's knowledge.}
    \label{fig:dialogue}
\end{figure}

%% file: sections/7_beyond.tex
\section{Beyond conversations}
\label{sec:beyond}

Beyond this modular setup we propose for modeling Mediators, we raise awareness of three further aspects to consider: How to evaluate explanation dialogues, when to allow customization of the explanation process, and how to train Mediators.

\subsection{Evaluating explanations}
In the HCXAI literature, previous works have employed evaluation measures beyond the estimation of the user's mental model, such as usefulness and satisfaction. This is analogous to the task of natural language generation, in that can show explanations to human subjects and ask them to rate and comment on them in various ways \cite{reiter-2019-natural}.
\cite{wiegreffe-marasovic-2021-review} identified two paradigms in this regard: Collect-and-Judge and Collect-and-Edit. While the former is about letting crowdworkers assess the quality of the (automatically or human-annotated) collected explanations, the latter necessitates annotators to edit the explanations to reduce annotation artefacts and biases and to improve quality control and linguistic variety.
This is echoed by \cite{arora-2022-explain-edit} who proposed explanation evaluation using iterative editing.
\cite{gonzalez-2021-interaction} evaluated explanations for reading comprehension models and showed that introducing multiple models of various quality and adversarial examples (which can be seen as another complementary type of explanations) can help to account for belief bias effects in human evaluation.

Moreover, we point the reader towards measurements for user trust and reliance on explanations and human-AI task performance \cite{mohseni-2021-multidisciplinary} as well as automated metrics for evaluating faithfulness \cite{hase-2020-leakage}.

Dialogue evaluation research has raised awareness of measuring flexibility and understanding among many other criteria \cite{mehri-2022-nsf}. There exist automated metrics based on NLP models for assessing the quality of dialogues, but their correlation with human judgments needs to be improved on.
These lines of research are disconnected from each other, which makes the task of evaluating explanation dialogues very challenging.
We argue that work on metrics specifically designed for evaluating explanation dialogues is just as important as implementational work.

\subsection{Customizing the explanation process}
For interactive explanations, \cite{sokol-flach-2020-fact-sheets} raised the need for controllability and customizability to suit a user's needs, e.g., through adjustable granularity. The complementary view of XAI methods proposed for Mediators allows for enhanced customizations of the dialogue and explanation narrative.

However, this sometimes requires offering additional settings adjustable via user interfaces, because they might not be easily communicated via natural language. Nevertheless, opening up the pandora's box of customization options ends up being a question of scalability, i.e. guaranteeing real-time responses \cite{miller-2019-explanation}.

%% file: sections/8_training.tex
\subsection{How to train Mediators}
\label{sec:train}

The most apparent open question about Mediators is how to train them. In the following, we will analyze existing datasets, give recommendations about future data collection and how previous frameworks utilized explanations for training models.

There is a distinct lack of natural language explanation datasets \cite{wiegreffe-marasovic-2021-review} covering explanation dialogues,
complicating the transfer to domain-specific use cases.
% Closest comparison / Data collection / Human-human dialogues
\cite{attari-2019-explainability} presented a study on how to collect data from human-human dialogues that can be used to train systems akin to Mediators.
\cite{madumal-2019-grounded} analyzed explanation dialogue transcripts and identified key components of such interactions.
\cite{weitz-2021-fault} analyzed how mental models are formed by users playing a collaborative puzzle game including an explanatory dialogue system.
We recommend going for this type of data if the Mediator's main responsibility is to hold dialogues that are as natural as possible and to understand questions that might be outside the scope that we presented in Tab.~\ref{tab:sst} and Tab.~\ref{tab:follow-up}.

% Information-seeking dialogue
Another close comparison are information-seeking dialogue or question answering datasets \cite{choi-2018-quac,saeidi-2018-interpretation,penha-2019-mantis,qu-2018-msdialog,feng-2020-doc2dial,dai-2022-dialog-inpainting}
% li-2020-molweni : Probably not relevant, because multi-party setup and Wh-questions are not transferable to our use cases
which already are known to be notoriously hard to create \cite{rogers-2021-explosion}.
We find, however, that such datasets are more concerned with covering a range of wh-questions instead of explanations (``Why...?'') \cite{wu-2022-qaconv}.

Based on our findings while reviewing the aforementioned datasets, we propose desiderata for explanation dialogue datasets:
At the minimum, they should include user utterances and feedback for post-dialogue user assessment.
However, to navigate the narrow path towards such richly annotated data, practitioners might have to invest a lot of effort in terms of implementation, costs and time to generate atomic explanations that humans can give feedback on, e.g. via Likert-scale ratings.

To which degree information-seeking question answering or human-human dialogue datasets can serve as training data for Mediators, depends on the kind of use case and goal.
For first test runs, they should be aware of the implementational effort and lack of control over the Mediator's output and favor Wizard-of-Oz studies \cite{sokol-flach-2020-one}.
We point the reader towards investigative work on the role of explanation data for training NLP models \cite{hase-bansal-2022-roles,hartmann-sonntag-2022-survey}.

%% file: sections/9_related-work.tex
\section{Related Work}
While \cite{miller-2019-explanation} is the main work advocating for conversational explanations, the recent work of \cite{lakkaraju-2022-rethinking} supported these theoretical foundations with a study where they interviewed domain experts about their needs and desires for such explanations. 
We see \cite{nobani-2021-conversational} as the (proposed) framework that is closest to a Mediator. However, they have yet to present empirical evidence and tie it to an actual use case. Our work is more comprehensive in comparison and connects the dots between the communities of HCXAI and NLP research.

Regarding Mediators in practice, we draw a connection towards applications which allow users to explore NLP models interactively \cite{tenney-2020-lit,strobelt-2021-lmdiff,strobelt-2022-genni,lee-2022-coauthor,perez-2022-red-teaming}. Although not all of their functional features might translate to conversational setups, we expect a dialogue-based interaction on NLP use cases (e.g., next-word prediction, summarization, story generation, question answering) to elicit more useful insights for all parties involved: The user is more engaged in a dialogue with conversational agents and the model can be trained on more elaborate responses.

%% file: sections/10_conclusion.tex
\section{Conclusion}

We engineered a blueprint of Mediators, conversational agents explaining the behavior of neural models in an interactive fashion.
We summarized the desiderata that HCXAI research put forward for dialogue-based explanations and highlighted that the current state of research in NLP has yet to catch up and address the gaps and pitfalls.
We recommended employing search methods in a complementary view of explanations and focussing on user expectations by keeping track of their mental models via rigorous, continuous evaluation.
We hope that this position paper inspires data collection and implementational work in Mediators for model behavior. 

%% file: sections/acknowledgments.tex
\section*{Acknowledgments}

We would like to thank Mareike Hartmann for extensive and fruitful discussions and Jan Nehring and Aljoscha Burchardt as well as the anonymous reviewers at the IJCAI 2022 Workshop on Explainable Artificial Intelligence for their valuable feedback. This work has been supported by the German Federal Ministry of Education and Research as part of the project XAINES (01IW20005).